\documentclass[10pt,twocolumn,letterpaper]{article}
\usepackage[utf8]{inputenc}

\usepackage{times}
\usepackage{epsfig}
\usepackage{graphicx}
\usepackage{amsmath}
\usepackage{amssymb}
\usepackage{xcolor}
\usepackage[ruled,vlined,onelanguage]{algorithm2e}
\usepackage{setspace}
\usepackage{collcell}
\usepackage{xr}
\usepackage{multirow}
\usepackage{subcaption}
\usepackage{mathtools}
\usepackage{colortbl,dcolumn}

\DeclarePairedDelimiterX{\infdivx}[2]{(}{)}{%
  #1\;\delimsize\|\;#2%
}

\SetKwInput{kwArgs}{Args}

\DeclareMathOperator{\cummax}{cummax}
\newcommand{\registered}{\textsuperscript{\tiny\textregistered}}
\newcommand{\etal}{\textit{et al.}}
\renewcommand{\eqref}[1]{\hyperref[#1]{Eq.\ \ref*{#1}}}
\newcommand{\figref}[1]{\hyperref[#1]{Fig.\ \ref*{#1}}}
\newcommand{\tabref}[1]{\hyperref[#1]{Table\ \ref*{#1}}}
\newcolumntype{C}[1]{>{\centering}m{#1}}

\newcommand*{\MinNumber}{0.0}%
\newcommand*{\MidNumber}{0.5} %
\newcommand*{\MaxNumber}{1.0}%

\newcommand{\ApplyGradient}[1]{%
        \ifdim #1 pt > \MidNumber pt
            \pgfmathsetmacro{\PercentColor}{max(min(100.0*(#1 - \MidNumber)/(\MaxNumber-\MidNumber),100.0),0.00)} %
            \hspace{-0.33em}\colorbox{green!\PercentColor!yellow}{#1}
        \else
            \pgfmathsetmacro{\PercentColor}{max(min(100.0*(\MidNumber - #1)/(\MidNumber-\MinNumber),100.0),0.00)} %
            \hspace{-0.33em}\colorbox{red!\PercentColor!yellow}{#1}
        \fi
}

\newcolumntype{R}{>{\collectcell\ApplyGradient}c<{\endcollectcell}}

\begin{document}

\title{Spectral Metric for Dataset Complexity Assessment}

\author{Frédéric Branchaud-Charron$^1$, Andrew Achkar$^2$, Pierre-Marc Jodoin $^1$\\
 $^1$ Université de Sherbrooke, $^2$ Miovision Inc.\\
{\tt\small $^1$\{frederic.branchaud-charron, pierre-marc.jodoin\}@usherbrooke.ca}, {\tt\small $^2$aachkar@miovision.com}
\thanks{This work is supported by the FQRNT B1 208594 and Mitacs acceleration IT08995.}
}

\maketitle

\begin{abstract}
In this paper, we propose a new measure to gauge the complexity of image classification problems.  Given an annotated image dataset, our method computes a complexity measure called the {\em cumulative spectral gradient} (CSG) which strongly correlates with the test accuracy of convolutional neural networks (CNN). The CSG measure is derived from the probabilistic divergence between classes in a spectral clustering framework. We show that this metric correlates with the overall separability of the dataset and thus its inherent complexity. As will be shown, our metric can be used for dataset reduction, to assess which classes are more difficult to disentangle, and approximate the accuracy one could expect to get with a CNN.  Results obtained on 11 datasets and three CNN models reveal that our method is more accurate and faster than previous complexity measures. \vspace{-0.4cm}
\end{abstract}

\section{Introduction}

The number of image-based datasets designed to train deep convolutional neural networks (CNN) have been on the rise in the past few years~\cite{bulatovNotMNIST,coates2011analysis,krizhevsky2009learning,lecun2010mnist,Luo18,netzer2011reading,yang2015large}.  One reason for this is the indisputable efficiency of CNNs at classifying image data~\cite{chollet2016xception,he2016deep,krizhevsky2009learning,simonyan2014very}.

A common challenge that arises when building a new image dataset for training a CNN is to identify how challenging the classification problem is, which classes are the most difficult to disentangle, and correspondingly what is the minimum dataset size required to train a CNN. As of today, there is no standard framework to make such determinations. The common way to assess the complexity of an image dataset is by training, finetuning and comparing results from several CNNs, the test accuracy being the usual measure for complexity.  However, this procedure is time consuming and, most importantly, requires a fully-annotated dataset which is not available when in the process of building it.

Unfortunately, one cannot predict the accuracy of a CNN by only looking at its architecture. As mentioned by Zhang \etal~\cite{Zhang17} in their attempt to understand why deep neural nets generalize well, deep neural networks can easily fit with zero training error on any input data, including pure random noise.  This underlines the sole ability of CNNs to project any input data into a linearly separable space (and thus have a zero training error) while sometimes having poor generalization abilities.  Their conclusion is that the structure of a neural net, its hyperparameters, its depth, and its optimizer cannot be used alone to predict its generalization capabilities.

Assessing the complexity of a classification problem may instead start from the analysis of the data at hand with the goal of deriving useful complexity measures ({\em c-measures})~\cite{Anwar14,Baumgartner06,Duin06,ho2002complexity,Sotoca06}.  
The goal of c-measures is to assess how entangled classes are assuming that datasets with overlapping classes are more difficult to analyze than those with well separated classes.  C-measures have been shown effective for a number of applications such as classifier selection~\cite{Brun18}, automatic noise-filtering adjustment~\cite{Saez13}, dataset reduction~\cite{Leyva15}, and hyperparameter tuning~\cite{Mantovani15}.

Unfortunately, existing c-measures have not been designed for large image datasets used to train deep neural networks.  While some c-measures assume that classes are linearly separable in their original feature space~\cite{ho2002complexity}, others work only for two-class problems~\cite{Anwar14,Baumgartner06,hoiem2012diagnosing}. Also, some c-measures are prohibitively slow and memory expensive as they require the analysis of matrices whose size is in the order to the number of training samples and/or the feature dimension size ~\cite{Baumgartner06,Duin06}.

Another important limitation with existing c-measures is the fact that they process raw input data.  While this was shown valid for some classification problems~\cite{Brun18}, it is ill suited for deep neural nets since their learning procedure allows them to project input data onto a different and more easily separable space.

In this paper, we present a novel c-measure adapted  to modern image classification problems.  Instead of processing raw input data like previous approaches, our method first projects the input images onto a lower-dimensional latent space.  This allows to analyze data whose features are better adapted to what CNNs learn.
Our method then estimates pairwise class overlap with a Monte-Carlo method which leads to an inter-class similarity matrix.  Following the spectral clustering theory, we compute a $K \times K$ Laplacian matrix where $K$ is the number of classes.  Finally, the spectrum of this matrix is used to derived our {\em cumulative spectral gradient} (CSG) c-measure.  

The main advantages of our proposed c-measure are as follows :
\begin{enumerate}
\item It naturally scales with the number of classes and the number of images in the dataset; \vspace{-0.2cm}
\item Our metric is fast to compute and does not require the computation of prohibitively large matrices; \vspace{-0.2cm}
\item It has no prior assumption on the distribution of the data; \vspace{-0.2cm}
\item It gives a strong insight on which classes are easily separable and those that are entangled; \vspace{-0.2cm}
\item The metric is highly correlated with CNN generalization capabilities; \vspace{-0.2cm}
\item It can be easily used for dataset reduction.
\end{enumerate}

\section{Previous works}

The goal of a c-measure is to characterize the difficulty of a classification problem.  While several c-measures have been proposed in the past, those by Ho and Basu are by far the most widely used~\cite{ho2002complexity}. They proposed 12 different descriptors called {\em F1, F2, F3, L1, L2, L3, N1, N2, N3, N4, T1,} and {\em T2}. F1 is a Fisher's Discriminant Ratio, F2 measures the inter-class overlap, and F3 is the largest fraction of points one can correctly classify with a stump decision function.  L1, L2 and L3 measures the linear separability of the data, while N1, N2, N3 and N4 are nearest neighbor measures which estimate the inter-class overlap.  As for T1, it measures the total number of hyperspheres one can fit into the feature space of a class and T2 is the ratio between the total number of training samples $N$ divided by the dimensionality of the data $d$. 

While the c-measures by Ho and Basu have been shown effective for small non-image datasets~\cite{Brun18}, those metrics are less suited to analyze large and complex image datasets.  For example, F1, F2, F3, L1, L2 and L3 assumes the data is linearly separable which is an over-simplistic assumption when considering modern image datasets. F1 requires the computation of $d\times d$ matrices which is problematic memory wise for large $d$ (i.e. for medium to large images) and F3 measures the linear separability of each class by accounting for each feature independently which is prohibitively slow when both $N$ and $d$ are large.  T1 is also prohibitively slow as $N$ gets large since it requires to grow an hypersphere around each data point and T2 is not a good complexity predictor as will be shown in the results section.  

Although Ho and Basu's metrics were designed for two-class problems, some researchers generalize it to more than two classes by averaging measures obtained between all possible pair of classes~\cite{orriols2010documentation,Sotoca06}.  Also, although recent generalizations of the Ho-Basu c-measures have been proposed~\cite{Anwar14,Cummins11,Sotoca06}, none addresses explicitly the problem of classifying large image datasets.

Other c-measures have been proposed. For example, Baumgartner and Somorjai~\cite{Baumgartner06} proposed a metric adapted to small biomedical datasets with high dimensionality data.  Unfortunately, their c-measures are for two-class problems, assume that the data is linearly separable and require the decomposition of $N\times d$ matrices which is only tractable when $N$ and $d$ are small.  Duin and Pekalska~\cite{Duin06} quantify the complexity of a dataset with metrics derived from a dissimilarity matrix of size $N \times M$ where $N$ is the training set size and $M$ is the number of “representation” vectors randomly sampled from the training set.   They report results on several datasets including two image datasets which contains 2000 or less black and white digits. The authors used the Euclidean distance to measure the similarity between two images, a metric that does not generalize well to real-world images~\cite{Wang05}.  

Like we do, some methods build a graph from the dataset to characterize the intra and inter-class relationships~\cite{Garcia15,morais2013complex}. This type of method requires building a $N \times N$ distance matrix which is problematic memory wise for large datasets. For example, the Hub score by \cite{Lorena18} requires to compute $A^TA$ where $A$ is a $N \times N$ adjacency matrix.  

To our knowledge, Li \etal~\cite{li2018measuring} are the only ones who proposed a c-measure applied specifically to modern image datasets and deep neural networks.  They called their measure the {\em Intrinsic Dimension} which is the minimum number of neurons a model needs to reach its best performances. They show that adding more neurons past the Intrinsic Dimension does not improve test accuracy. Unfortunately, as opposed to what we seek to do, their measure requires multiple training of image classification CNNs through a grid-search approach which is slow and tedious.  More details on c-measures can be found in the recent survey paper by \cite{Lorena18}.

\section{Proposed Method}

\subsection{Class overlap}
At the core of our c-measure is the notion of class overlap.  Let $x$ be an input image and $\phi(x)\in I\!\!R^d$ an embedding for that image.  As will be  discussed later, $\phi$ can be any function that projects $x$ to a new dimensional space where images with similar content are close together and the other ones further away.  The overlap between two classes $C_i$ and $C_j$ refers to the overall area in the feature space for which $P(\phi(x_k)|C_i)>P(\phi(x_k)|C_j)$ when $\phi(x_k)$ is a member of class $C_j$.  Class overlap can thus be formulated as~\cite{Nowakowska14}:
\begin{eqnarray}
\label{eq:overlap}
\int_{I\!\!R^d}\min \left ( P(\phi(x)|C_i),P(\phi(x)|C_j) \right )d\phi(x).
\end{eqnarray}
Unfortunately, the direct calculation of this integral is prohibitively complicated for non-parametric distributions and when $d$ (the dimensionality of the embedded space) is large.  Since the overlap between two classes is related to the similarity of their distributions, one may instead use a probability distribution distance function such as the Kullback-Leibler divergence or the Kolmogorov-Smirnov test as a surrogate for Eq.(\ref{eq:overlap}).  One such  function is the probability product kernel of Jebara \etal~\cite{jebara2004probability} :
\begin{eqnarray}
\label{eq:jabara}
\int_{I\!\!R^d}  P(\phi(x)|C_i)^\rho P(\phi(x)|C_j)^\rho d\phi(x)
\end{eqnarray}
which is a generalization of the Bhattacharyya kernel (and the Hellinger distance) when $\rho=1/2$.  While computing Eq.(\ref{eq:jabara}) is as complex as computing Eq.(\ref{eq:overlap}) for an arbitrary value of $\rho$, simplification occurs when $\rho=1$.  In that case, the kernel becomes the inner-product between the two distributions $\int_{I\!\!R^d}  P(\phi(x)|C_i)P(\phi(x)|C_j) d\phi(x)$  which is the expectation of one distribution under the other : $E_{P(\phi(x)|C_i)}[P(\phi(x)|C_j)]$ or $E_{P(\phi(x)|C_j)}[P(\phi(x)|C_i)]$.  
 
 Formulating the inter-class divergence as an expectation function allows one to use Monte-Carlo to approximate it: 
\begin{eqnarray} \label{eq:mce}
E_{P(\phi(x)|C_i)}[P(\phi(x)|C_j)] \approx \frac{1}{M}\sum_{m=1}^M P(\phi(x_m)|C_j)
\end{eqnarray}
where $\{\phi(x_1), ..., \phi(x_M)\}$ are $M$ samples {\em i.i.d.} from $P(\phi(x)|C_i)$.  One can thus approximate the divergences between two class distributions by averaging the probability of $M$ samples of class $C_i$ to be in class $C_j$ or vice versa.  
Computing inter-class divergences leads to a $K\times K$ similarity matrix $\cal S$ where $K$ is the total number of classes and ${\cal S}_{ij}$ is the Monte-Carlo approximation of  the divergence between $C_i$ and $C_j$. 


Since the underlying model of $P(\phi(x_m)|C_j)$ is {\em a priori} unknown, we approximate it with a K-nearest estimator:
\begin{eqnarray}
\label{eq:knn}
p(\phi(x)\mid C_j) =  \frac{K_{C_j}}{MV} 
\end{eqnarray}
where V is the volume of the hypercube surrounding the $k$ closest  samples  to $\phi(x)$ in class $C_j$, $M$ is the total number of samples selected in class $C_j$ and $K_{C_j}$ is the number of neighbors around $\phi(x)$ of class $C_j$.

\subsection{Spectral Clustering}
The $K\times K$ similarity matrix $\cal S$ embodies the overall complexity of a dataset by means of class overlap.  Our goal is to extract a measure from $\cal S$  that would summarize the complexity of that dataset.  For that, we rely on the spectral clustering theory~\cite{von2007tutorial} that we briefly review in this section.  

Let $G$ be an undirected similarity graph $G=(V,E)$ where $V$ is a set of nodes connected by edges $E$.  An edge $E_{ij}$ is an arc connecting two nodes $i$ and $j$ and whose weight $w_{ij}\geq 0$ encodes how close these two nodes are.  A weight of 0 implies no connection between $i$ and $j$ whereas a large weight implies strong similarity (in our case, a weight of 1 implies that $i$ and $j$ are identical).  The weight of all edges are put in a $n\times n$ adjacency matrix $W$  where $n$ is the total number of nodes.  Note that $W$ is symmetric and positive semi-definite due to the undirected nature of the graph which implies that $w_{ij}=w_{ji}$.

The goal of spectral clustering is to partition $G$ into subgraphs such that the edges between the subgraphs have minimum weight. A set of subgraphs $\{ G_1,...,G_l\}$ is valid when  $G_i \cap G_j=\emptyset,$ $\forall i\not=j$ and $G_1\cup ... \cup G_l=G$.  An optimal partition of $G$ is one for which the cut has minimum cost : $costCut(G_1,...,G_l)=\sum w_{ij}$ for $i$ and $j$ in different subgraphs.  

Spectral clustering provides an elegant framework to recover the subgraphs with minimum cut.  It starts with a Laplacian matrix whose simplest form is $L = D - W$ where $D$ is a degree matrix $D_i = \sum_j w_{i,j}$.  $L$ is symmetric and positive semi-definite, it contains $n$ eigenvalues $\{ \lambda_0, ..., \lambda_{n-1} \}$ that are real and non negative with $\lambda_{0} = 0$ and $\lambda_{i+1}\geq \lambda_{i}$. This set of eigenvalues is called the \textit{spectrum} of $L$.  Interestingly, the $n$ eigenvectors associated to the eigenvalues can be seen as indicator vectors that one can use to cut the graph.  Also, the magnitude of their associated eigenvalues is related to the cost of their cut~\cite{Mohar1997}.  As such, the eigenvectors associated to the lowest eigenvalues are those associated to the partitions of minimum cost. 

\subsection{Inter-class adjacency matrix}

We formulate our c-measure within the spectral clustering framework for which each node is a class index.  In our case, $W$ and $L$ are $K \times K$ matrices where $K$ is the total number of classes.  As such, the weight $w_{i,j}$ is the distance between the likelihood distributions of classes $C_i$ and $C_j$.  Thus, a simple dataset for which each pair of classes has little overlap would produce a sparse Laplacian matrix $L$ whose spectrum  contains small eigenvalues. On the other hand, a more complex dataset with stronger class overlap would lead to a spectrum with larger eigenvalues.  

Since the similarity matrix $\cal S$ was obtained with a Monte-Carlo approximation of the Jebara kernel, it is not symmetric and thus cannot be used as an adjacency matrix $W$.  Instead, we consider each column ${\cal S}_i$ as a signature vector of each class $i$ so two classes with similar likelihood distributions would also have a similar signature vector ${\cal S}_i$ and vice versa.  We then compute $W$ following the Bray-Curtis distance function~\cite{greenacre2013measures}: 
\begin{eqnarray}\label{eq:wij}
w_{ij} = 1- \frac{\sum_k^K{|{\cal S}_{ik}-{\cal S}_{jk}|}}{\sum_k^K{|{\cal S}_{ik}+{\cal S}_{jk}|}}.
\end{eqnarray}
This equation implies that $w_{ij}=0$ when the distributions of classes $i$ and $j$ do not overlap and $w_{ij}=1$ when the distributions are identical.

\subsection{Runtime improvement}

Computing the adjacency matrix $W$ with the  Bray-Curtis function as well as the Monte-Carlo method (Eq.(\ref{eq:mce})) is 40 times faster than with a naive implementation (Eq.(\ref{eq:jabara})) for a K=10 class problem. This  explains why our method is fast and gets good results even with a small number of samples.  We came to that number as follows.

First, let us mention that the most computationally intensive operation is the point-wise estimation of a probabilistic distribution function (pdf) $P(\phi(x)|C)$.  
Since computing Eq.(\ref{eq:jabara}) requires $M$ estimations of $P(\phi(x)|C_j)$, the $K\times K$ similarity matrix $S$ requires a total of $K^2 \times M$ pdf estimations,  where $K$ is the number of classes and $M$ the number of samples.  Also, since Eq.(\ref{eq:wij}) requires no additional pdf estimation, our method requires a total of $K^2 \times M$ pdf estimations to compute the adjacency matrix $W$.

However, since the Bray-Curtis distance function combines two $R^K$ vectors $S_i$ and $S_j$, it incorporates the statistics of $2\times K \times M$ samples at each entry $w_{ij}$ of $W$.  If the same number of samples were to be used by a naive implementation, i.e. that $w_{ij}$ was to be computed with $2\times K \times M$ samples and Eq.(\ref{eq:jabara}), the computation of $W$ would require a total of $4\times K^3\times M$ pdf estimations, i.e. $4\times K$ more pdf estimations than for our method.  From there we conclude that our method is 40 times faster than a naive implementation when $K=10$.  Please note that these findings are in line with empirical results.


\begin{figure}[tp]
    
    \centering
    \hspace{-2cm}
	\begin{minipage}[c]{0.32\textwidth}
		\centering    
        \includegraphics[width=\textwidth]{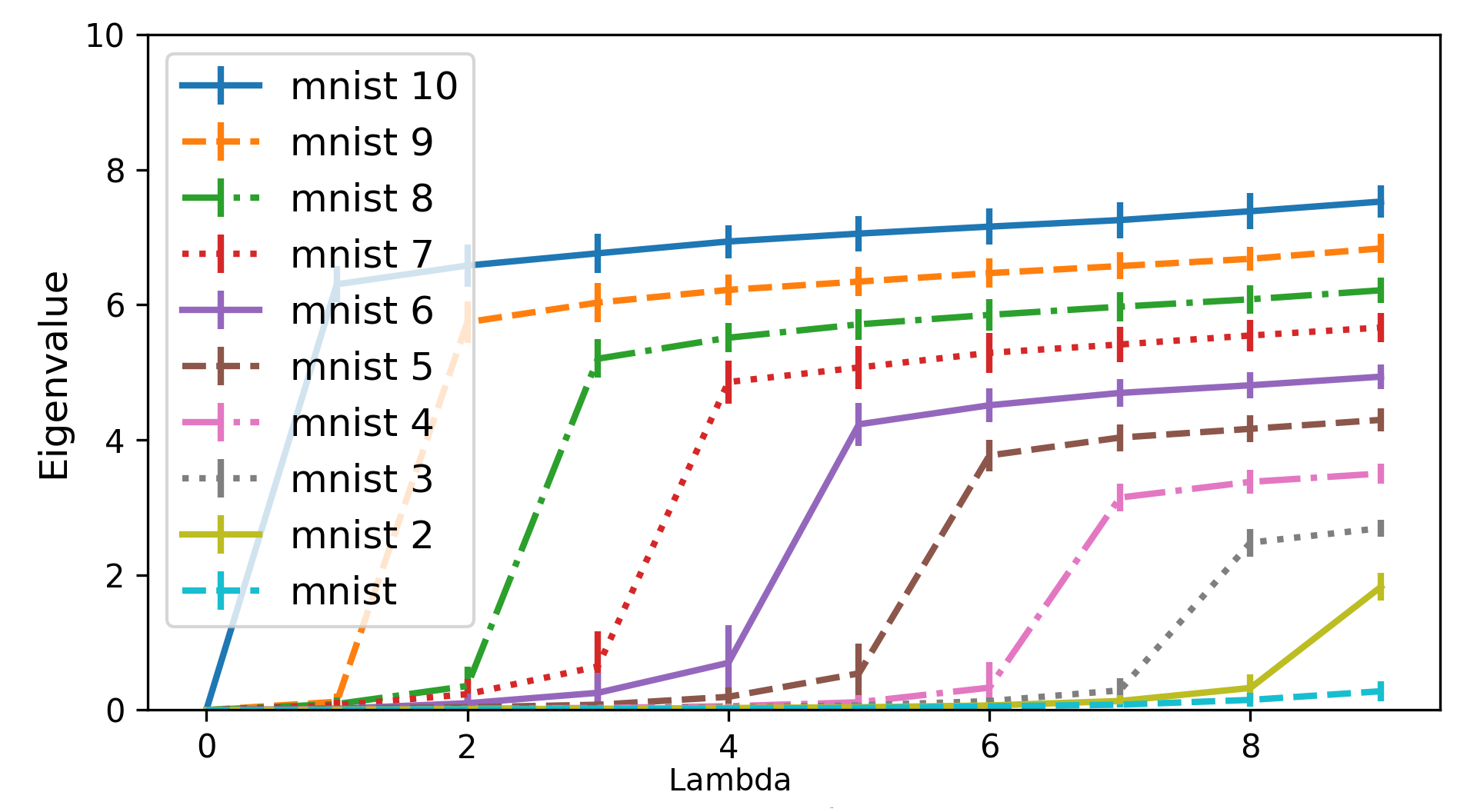}
        \label{fig:mnist_eigens}
     \end{minipage}
    ~ 
   \begin{minipage}[c]{0.05\textwidth}
   	  \centering
      \vspace{-0.5cm}
      \scalebox{0.7}{
      \begin{tabular}{ccc}
        Dataset & CSG & E.R.\\
        \hline
        mnist 10 & 5.51 & 0.91 \\
        mnist 9 & 5.04 & 0.78\\
        mnist 8 & 4.53 & 0.69 \\
        mnist 7 & 3.79 & 0.61 \\
        mnist 6 & 3.31 & 0.51 \\
        mnist 5 & 2.70 & 0.39 \\
        mnist 4 & 2.16 & 0.30 \\
        mnist 3 & 1.52 & 0.18 \\
        mnist 2 & 0.84 & 0.13\\
        mnist & 0.12  & 0.01 \\
      \end{tabular}}

   \end{minipage}
    \caption{[Left] Spectrum of ten noisy versions of MNIST and [right] our CSG c-measure with the error rate (E.R.) of an AlexNet CNN (figure best viewed in color).\vspace{-0.3cm}}\label{fig:mnistres}
\end{figure}

\subsection{The CSG complexity measure}

As mentioned before, a dataset with a low eigenvalue spectrum indicates  a low inter-class overlap and thus easily separable classes.  To illustrate this, we put in Fig.~\ref{fig:mnistres} the spectrum of the MNIST dataset (the bottommost cyan curve) which we obtained by processing raw images.  Since MNIST contains 10 classes, its spectrum contains 10 eigenvalues.  Being a simple dataset, MNIST's spectrum contains mostly near-zero values.  We then randomly swap elements between classes to force their distribution to strongly overlap, making this noisy version of MNIST more complex.  We first swap elements between two classes (MNIST 2), then between three classes  (MNIST 3) all the way to 10 classes  (MNIST 10).  As one can see, these noisy versions of MNIST lead to a gentle progression of the spectrum profiles.  The more entangled classes are, the larger the eigenvalues are. Also, the sooner a strong spectrum gradient occurs ($\lambda_{i+1}-\lambda_i$) the more difficult the dataset is (this gradient discontinuity is also called the {\em eigengap} in the spectral clustering literature~\cite{von2007tutorial}).  

The overall complexity of the dataset is thus related to  the area under the spectrum curve as well as the position of the eigengap.  To account for both observations, we first normalize the eigengap by its horizontal position:

\begin{eqnarray}\label{eq:deltalambda}
\Delta \widetilde{\lambda_i} = \frac{\lambda_{i+1} - \lambda_i}{K - i}.
\end{eqnarray}
%
The normalization by $K-i$ is at the core of our metric. Depending on where the largest eigengap occurs, its maximum value can only be of $K-i$. The difficulty of cutting the graph is thus related to the position of the largest eigengap.
Our c-measure (the {\em cumulative spectral gradient} (CSG)) is the cumulative maximum (cummax) of the normalized eigengaps :
\begin{eqnarray}\label{eq:csg}
CSG = \sum_i \cummax(\Delta \widetilde{\lambda})_i.
\end{eqnarray}
With a cummax, between two spectrums with the same area under the curve, our CSG measure will be larger for the one with the left-most eigengap.  The CSG values for the noisy MNIST datasets are shown on the right of Fig.~\ref{fig:mnistres} along side with the test error rate obtained with an AlexNet CNN~\cite{krizhevsky2012imagenet}.  As can be seen, our CSG c-measure is heavily correlated to the complexity of the datasets. 

Our method is summarized in Algorithm ~\ref{alg:algo1}. 

\begin{algorithm}[h]
\setstretch{1.1} 
\SetAlgoLined
 \KwData{Dataset=$\{(\phi(x_1),t_1),...,(\phi(x_N),t_N)\}$}
 \kwArgs{M, k}
 \KwResult{CSG score}
 Compute inter-class similarity matrix $\cal S$ with Eq.(\ref{eq:mce}) and (\ref{eq:knn}) $\forall$ pair of classes $C_i, C_j$\\
 Compute W (Eq.(\ref{eq:wij})) \\
 $L \leftarrow D-W $ \\
 $\{\lambda_1,...,\lambda_K\} \leftarrow \mbox{EigenValues}(L)$ \\
 Compute CSG (Eq.(\ref{eq:csg})) \\
 \Return CSG 
 
 \caption{The CSG c-measure algorithm.}
 \label{alg:algo1}
\end{algorithm}

\vspace{-0.2cm}
\section{Results}
\label{sec:res}

\subsection{Embeddings}
As mentioned before, the input images $x$ are projected to an embedding space with a function $\phi(x)$.  In this paper, we tested four projection functions : 
\begin{enumerate}
\item {\em Raw}; the identity function $\phi(x)=x$ ; \vspace{-0.2cm}
\item {\em t-SNE}; the t-SNE function~\cite{van2008visualizing} which projects the raw input images down to a 2D space; \vspace{-0.2cm}
\item {\em CNN{\tiny $_{AE}$}}; the embedding of a 9-layer CNN-Autoencoder trained for 100 epochs; \vspace{-0.2cm}
\item {\em CNN{\tiny $_{AE}$} t-SNE}; the t-SNE function applied to the embedding of the CNN-autoencoder.
\end{enumerate}

\subsection{Datasets}

In order to gauge performance of our method, we used several image classification datasets of various difficulty levels.  Of those datasets, six contain 10 classes, one contains 11 classes and three contain two classes.  These datasets are summarized in Table~\ref{tab:datasets} and sorted according to the test error rate (E.R.) obtained with an AlexNet CNN~\cite{krizhevsky2012imagenet}. Note that we replaced the AlexNet local response norm with a batch-norm~\cite{ioffe2015batch}, trained it for 500 epochs on each dataset with a batch size of 32 and the SGD optimizer with the same parameters than in the original paper but without data augmentation. We used Keras~\cite{chollet2015keras}, Tensorflow~\cite{abadi2016tensorflow} and an Nvidia Titan X GPU.

The datasets are the well-known MNIST~\cite{lecun2010mnist} and  CIFAR10~\cite{krizhevsky2009learning}.  There is also notMNIST~\cite{bulatovNotMNIST}, a synthetic dataset of 18,724 letters made of unconventional fonts, and the Street View House Numbers (SVHN) dataset~\cite{netzer2011reading}, one of the most challenging digit classification datasets with 73,257 images of low resolution street numbers. 
We also use MioTCD~\cite{Luo18}, a large dataset of 648,959 vehicles pictured by traffic cameras with varying orientation angles, resolution, time of the day and weather conditions.  
STL-10~\cite{coates2011analysis} is a 10-class dataset similar to CIFAR-10 but with larger images ($96\times 96$ instead of $32\times 32$) and fewer training samples (5,000 instead of 50,000). SeeFood~\cite{becker2017seefood} is a two-class dataset ({\em Hot-dog} vs {\em No Hot-dog}) with 498 samples derived from the Food-101 dataset~\cite{bossard14food101}.  We also use the well-known Inria pedestrian dataset~\cite{dalal2005inria} containing 38,634 RGB images of pedestrians or not, and  Pulmo-X~\cite{jaeger2014two}, a two-class pulmonary chest X-Ray dataset for tuberculosis detection containing 662 images. Finally, CompCars~\cite{yang2015large} is a dataset containing 1,716 car categories of different makes and models. For our experiments, we selected the 10 makes with the highest count and resized the images to $128 \times 128$, giving us $500$ samples per class.

We followed the evaluation methodology specific to each dataset, i.e. we trained and tested the methods on the training and testing set provided with the datasets. For the two datasets without pre-determined train/test split (notMNIST and Pulmo-X) we made a 80-20 Train/Test split and kept the same class proportion.

\begin{table}[tp]
\center
  \begin{tabular}{|c|c|c|c|c|}
  \hline
  Datasets & E.R.& $K$ & $N$ & Content \\
  \hline
  MNIST    & 0.01 & 10 & 50k & Hand written digits \\
  MIO-TCD   & 0.03 & 11 & 649k & Traffic images\\
  notMNIST & 0.05 & 10 & 18.7k & Printed digits\\
  SVHN     & 0.08 & 10 & 73.3k & Printed digits \\
  Inria    & 0.10 &  2 & 3.6k & Pedestrians \\
  CIFAR10  & 0.12 & 10 & 50k & Various real images \\
  Pulmo-X  & 0.23 & 2  & 662 & Pulmonary X-Rays \\
  SeeFood  & 0.38 & 2  & 500 & Images of food \\
  STL-10   & 0.68 & 10 & 5k & Various real images \\
  CompCars & 0.70 & 10 & 6k & Pictures of cars \\
  
  \hline
  \end{tabular}
  \caption{Datasets used to validate our method with the test error rate (E.R.) of an AlexNet CNN~\cite{krizhevsky2012imagenet}, the number of classes $K$, the training set size $N$ and  a short summary.\vspace{-0.4cm}}
  \label{tab:datasets}
\end{table}

\vspace{-0.3cm}
\subsubsection{Hyper-parameters}
Our algorithm has two main hyper-parameters: $M$ the number of samples per class used by the Monte Carlo method in Eq.(\ref{eq:mce}) and $k$ the number of neighbors to compute the likelihood distribution of each class in Eq.(\ref{eq:knn}). In Table~\ref{tab:gridsearch}, we show the Pearson correlation score between our c-measure with the {\em CNN{\tiny $_{AE}$} t-SNE} embedding and the error rate of AlexNet on the six 10-class datasets (upper table) as well as the average processing time of our Algo 1 (lower table).  As one can see, the choice for $k$ and $M$ has little impact on the quality of the results (except for when $M$ is very small).  Also, while the runtime scales almost linearly with $M$, our method is still fast with timings below 3 seconds, even with $M=400$ samples per class. This shows that our method does not require a careful adjustment of its hyper-parameters. We found this as well for the other embeddings we tested. As such, we will use $M=100$ and $k=3$ for the remainder of this section.

\begin{table}[tp]
\centering
{\small
\begin{tabular}{cc|c|c|c|c|c|c|c}

& \multicolumn{1}{c}{ }& \multicolumn{6}{c}{$k$}  \\ \cline{3-8}
    & \multicolumn{1}{c}{} & \multicolumn{1}{c}{\textbf{1}} & \multicolumn{1}{c}{\textbf{3}} & \multicolumn{1}{c}{\textbf{5}} & \multicolumn{1}{c}{\textbf{7}} & \multicolumn{1}{c}{\textbf{9}} & \multicolumn{1}{c}{\textbf{11}} \\ 
\multirow{ 6}{*}{$M$}  & \textbf{2} &  0.81&  0.79&	 0.80&	 0.75&	 0.76&	 0.73 & \multirow{ 6}{*}{\rotatebox{270}{Pearson Corr.}}\\  
& \textbf{50} &  0.97&  0.97&	 0.97&	 0.96&	 0.96&	 0.97\\
&\textbf{100} &  0.97&	 0.97&	 0.98&	 0.98&	 0.97&	 0.97\\
&\textbf{200} &  0.98&	 0.98&	 0.98&	 0.98&	 0.98&	 0.98\\
&\textbf{300} &  0.97&	 0.97&	 0.97&	 0.98&	 0.97&	 0.98\\
&\textbf{400} &  0.97&	 0.97&	 0.97&	 0.97&	 0.97&	 0.97\\  \\

\multirow{ 6}{*}{$M$}& \textbf{2} & 0.02&	0.02&	0.02&	0.02&	0.02&	0.02 & \multirow{ 6}{*}{\rotatebox{270}{Timing (s)}}\\
& \textbf{50} &  0.30&	0.30&	0.29&	0.30&	0.29&	0.27\\
&\textbf{100} &  0.60&	0.61&	0.60&	0.60&	0.60&	0.60\\
&\textbf{200} &  1.22&	1.21&	1.23&	1.19&	1.20&	1.22\\
&\textbf{300} &  1.82&	1.82&	1.82&	1.78&	1.83&	1.79\\
&\textbf{400} &  2.42&	2.38&	2.42&	2.41&	2.39&	2.39\\ 
\end{tabular}}

\caption{Correlation values [upper table] and average processing times of Algo 1 in seconds [lower table] for various combinations of hyperparameters $M$ and $k$.}
\label{tab:gridsearch}
\end{table}

\subsection{Experimental results}


\begin{table}[tp]
\centering
{\small
\begin{tabular}{|c|ccc|}
\hline
c-measure &  Corr. & p-value & Time (s) \\
\hline
N4  & 0.069 & 0.896 & 3,744 \\
F3  & 0.267 & 0.610 & 3,924 \\
F1  & 0.501 & 0.311 & 72 \\
F2  & 0.422 & 0.405 & 72 \\
T1  & 0.357 & 0.487 & 36,108\\
T2  & 0.636 & 0.175 & 72 \\
N2  & 0.652 & 0.161 & 36,180 \\
F4  & 0.725 & 0.103 & 3,644 \\
N1  & 0.741 & 0.092 & 17,748 \\
N3  & 0.773 & 0.072 & 36,216 \\
\hline
$\max_i \lambda_i$ CNN{\tiny $_{AE}$} t-SNE & 0.88 & 0.02 & 0.3 (18,900)\\
$\sum_i \lambda_i$ CNN{\tiny $_{AE}$} t-SNE & 0.94 & $\leq$\textbf{0.01} & 0.3 (18,900)\\
\hline
CSG Raw & 0.696 & .125 & 50 (\textit{NA}) \\ 
CSG CNN{\tiny $_{AE}$} & 0.823 & .044 & 3.6 (13,300) \\
CSG t-SNE & 0.903 & .014 & 0.7 (6,084)\\
CSG CNN{\tiny $_{AE}$} t-SNE & \textbf{0.968} & $\leq$\textbf{0.01} & 0.3 (18,900) \\
\hline
\end{tabular}}
\caption{Correlation between the accuracy of AlexNet on 6 datasets and 10 c-measures by Ho-Basu~\cite{ho2002complexity} and ours methods with four embeddings, the associated p-value and processing time alongside the time to train the autoencoder in the parentheses (measured on CIFAR10).\vspace{-0.4cm}}
\label{tab:corr}
\end{table}

\subsubsection{Comparison with other c-measures}

We compared our method to the most widely implemented c-measures, i.e. those by Ho and Basu~\cite{ho2002complexity}.  We used the C++ DCol library provided by the authors~\cite{orriols2010documentation} and processed the six 10-class datasets. 
We thus followed the original methodology provided by the authors which implies no embedding. In addition, we tested two other metrics derived from the spectral theory: the maximum eigenvalue ($\max \Lambda$) and the area under the curve (AUC). These methods are known in the literature as being related to the similarity between nodes~\cite{Sorgun2013bounds}. These turn out to perform worse than our CSG metric. Results are reported in Table~\ref{tab:corr} together with our method with four embeddings.

\begin{figure}[tp]
    \centering
	\begin{minipage}[c]{0.49\textwidth}
		\centering    
        \includegraphics[width=0.9\textwidth]{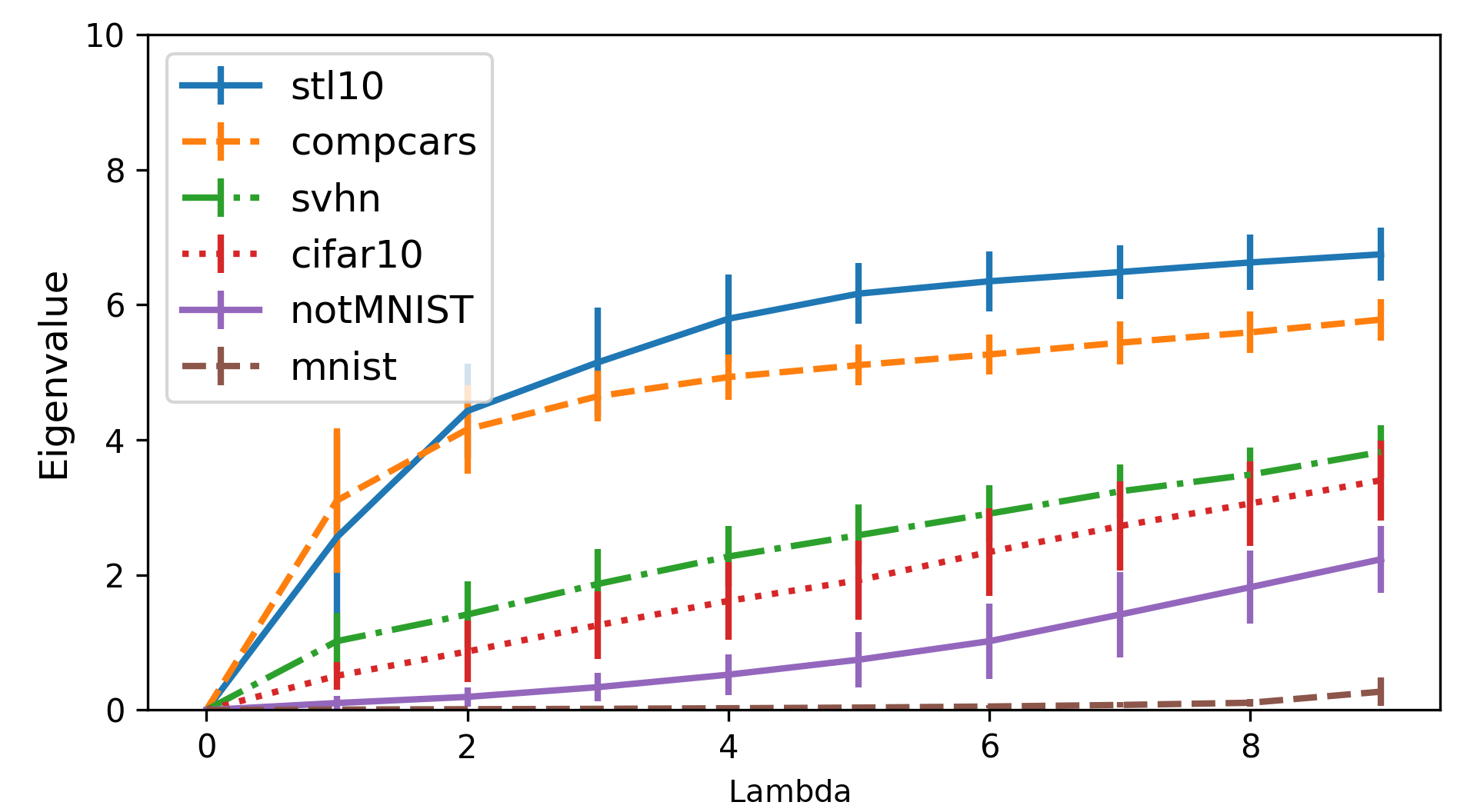}
     \end{minipage}
    ~ 
    \caption{Laplacian spectrum for the 10-class datasets.\vspace{-0.4cm}} \label{fig:spectrum_res}
    \label{fig:spectrum10datasets}
\end{figure}

The first column contains the Pearson correlation score between the error rate  by an AlexNet CNN and each c-measure.  As one can see, our method with the {\em CNN{\tiny $_{AE}$}}, {\em t-SNE} and {\em CNN{\tiny $_{AE}$} t-SNE} embeddings have a better correlation than any of the existing c-measures with a p-value below the $0.05$ bar.   The best embedding is CNN{\tiny $_{AE}$} t-SNE with a significance p-value below $0.01$.  To illustrate how this embedding correlates with the dataset complexity, we put in Fig.~\ref{fig:spectrum10datasets} its Laplacian spectrum for the six 10-class datasets.  As can be seen, the spectrum plots grow smoothly from the simplest dataset (MNIST), to slightly more complex datasets (notMNIST, CIFAR10 and SVHN) all the way to the most complex datasets (STL-10 and CompCars).  Note that we will use the CNN{\tiny $_{AE}$} t-SNE embedding for the remaining of this section.

As for processing time, our method is faster than the best c-measures F4, N1 and N3.  Note the value  on the left is the time to execute Algo. ~\ref{alg:algo1} whereas the value in parenthesis is the processing time to train a CNN{\tiny $_{AE}$} and/or run t-SNE.  Although that processing time is large (more than one hour) it is much faster than the previous best method N3.\footnote{The timings were computed on CIFAR10 using a Intel\registered Xeon\registered CPU E5-1620 and a NVIDIA TITAN X.} The performance of the t-SNE embedding is due to the fact that while t-SNE does not change the nearest neighbours, it does not preserve long-range distances which results in a less convoluted low-dimensional representation. In consequence, the approximation of the volume is better in this representation.

In Table~\ref{tab:multimethodsmultidatasets}, we provide our CSG c-measure with the test error rate of three CNN models as well as their Pearson correlation and p-value.  As can be seen, our c-measure correlates well not only with AlexNet, but also with more recent ResNet-50~\cite{he2016deep} and XceptionNet~\cite{chollet2016xception}.  Also, our correlation and p-value with CNN error rates is significantly better than the best existing c-measures~\cite{ho2002complexity} even when using our CNN t-SNE embedding. In fact, using embeddings seem detrimental to the overall performance of the existing c-measures. Results on all the existing c-measures with all the embeddings are available in the supplementary material.

\begin{table}[h]
\centering
  {\small
  \begin{tabular}{|c||c|C{0.92cm}|C{0.8cm}|c|}
  \hline
  \multicolumn{2}{|c|}{}&\multicolumn{3}{|c|}{Error rate}\\ \hline
  Datasets & CSG &$\!\!$AlexNet&$\!\!$ResNet-50& Xception \\ \hline
  CompCars & 2.93 & 0.70 & 0.88	& 0.86 \\
  STL-10    & 3.07 & 0.69 & 0.63 & 0.69 \\
  CIFAR10  & 1.00 & 0.18 & 0.19	& 0.06 \\
  SVHN     & 1.15 & 0.08 & 0.07 & 0.03 \\
  notMNIST & 0.72 & 0.05 & 0.04 & 0.03 \\
  MNIST    & 0.11 & 0.01 & 0.05 & 0.01 \\ \hline 
  \multicolumn{2}{|c|}{Method}&\multicolumn{3}{|c|}{Pearson correlation}\\ \hline
  \multicolumn{2}{|c|}{N3/CNN Corr} & 0.773 & 0.727 & 0.681 \\
  \multicolumn{2}{|c|}{N3/CNN p-val} & 0.054 & 0.102 & 0.136 \\ \hline
  \multicolumn{2}{|c|}{N3 {\footnotesize CNN t-SNE}/CNN Corr}  & 0.837 & 0.765 & 0.837 \\
  \multicolumn{2}{|c|}{N3 {\footnotesize CNN t-SNE}/CNN p-val} & 0.063 & 0.124 & 0.144 \\ \hline
  \multicolumn{2}{|c|}{CSG {\footnotesize CNN t-SNE}/CNN Corr} & \textbf{0.968} & \textbf{0.935} & \textbf{0.951} \\
  \multicolumn{2}{|c|}{CSG {\footnotesize CNN t-SNE}/CNN p-val} & \textbf{0.01} & \textbf{0.006} & \textbf{0.003} \\ \hline
  
  \end{tabular}}
  \caption{[Top] CSG c-measure alongside with test error rates for 3 CNN models on six datasets.[Bottom] Pearson correlation and p-value between the CNN error rates and the N2 and N3 Ho-Basu our CSG c-measure ~\cite{ho2002complexity} and CNN.}
  \label{tab:multimethodsmultidatasets}
\end{table}

We also tested our method on two-class image classification problems. We used the Inria, SeeFood, and PulmoX datasets as well as the deer-dog subset of CIFAR10.  Results reported in Table~\ref{tab:multimethodsmulti2Cdatasets} show that our method correlates well with the CNN models, especially AlexNet.  Our correlation scores are also better than those of the best c-measure of Ho-Basu (although by a slight margin) although it was specifically designed for two-class problems.

\begin{table}[h]
\centering
  {\small
  \begin{tabular}{|c||c|C{0.92cm}|C{0.8cm}|c|}
  \hline
  Datasets & CSG &$\!\!$AlexNet&$\!\!$ResNet-50& Xception \\ \hline
  SeeFood  & 0.95 & 0.38 & 0.34 & 0.21 \\ 
  PulmoX   & 0.55 & 0.23 & 0.16	& 0.11 \\
  deer-dog & 0.39 & 0.20 & 0.02 & 0.02 \\
  Inria    & 0.32 & 0.10 & 0.07	& 0.03 \\ \hline \hline
  \multicolumn{2}{|c|}{N3 Ho-Basu/CNN Corr} & 0.976 & 0.852 & 0.862 \\
  \multicolumn{2}{|c|}{N3 Ho-Basu/CNN p-val} & 0.01 & 0.148 & 0.138 \\ \hline
  \multicolumn{2}{|c|}{CSG/CNN Corr} & {\bf 0.995} & {\bf 0.860} & {\bf 0.887} \\
  \multicolumn{2}{|c|}{CSG/CNN p-val} & {\bf 0.006} & {\bf 0.130} & {\bf 0.113} \\ \hline
  
  \end{tabular}}
  \caption{[Top] CSG c-measure alongside with test error rates for 3 CNN models on four 10-class datasets.[Bottom] Pearson correlation and p-value between ours methods and CNN and between the N3 c-measure~\cite{ho2002complexity} and CNN.}
  \label{tab:multimethodsmulti2Cdatasets}
\end{table}

\subsubsection{Dataset reduction}
Dataset reduction (also known as {\em instance selection}~\cite{Leyva15}) consists in reducing as much as possible the number of elements in a dataset without losing trained CNN accuracy.  One way of doing so is by iteratively removing elements from the dataset up to a point where the CSG measure increases sharply.

\begin{figure}[tp]
\centering
\includegraphics[width=0.8\linewidth]{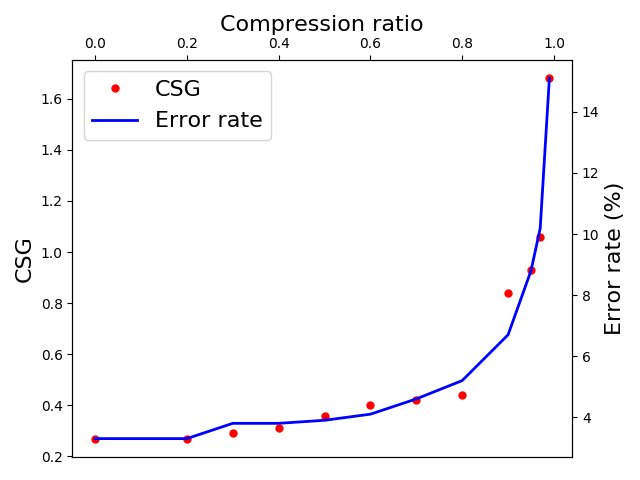}
\caption{Our c-measure and AlexNet accuracy obtained while reducing the size of the MioTCD dataset.}
\label{fig:miotcdreduction}
\end{figure}

We first tested our method on the MIO-TCD dataset~\cite{Luo18}, a large dataset used for a 2017 CVPR challenge and for which CNN methods got accuracies of up to 98\%.  Such high accuracies suggest  that the dataset is overcomplete and could be reduced without affecting much the CNN accuracy.  Results for various reduction ratios are shown in Figure~\ref{fig:miotcdreduction}.  As one can see, the CSG (red dots) stays roughly unchanged for reduction ratios below $80\%$ but then increases sharply after that.  This is inline with the AlexNet test error rate (blue line) although it took less than 5 minutes to produce the CSG measures and 5 days the AlexNet results. We used the same CNN{\tiny $_{AE}$} embedding for all ratios.  We got a Pearson correlation of 0.956 between our CSG dots and the error rate values shown in Figure~\ref{fig:miotcdreduction}.

\begin{figure*}[htp]
\centering
\includegraphics[width=0.3\textwidth]{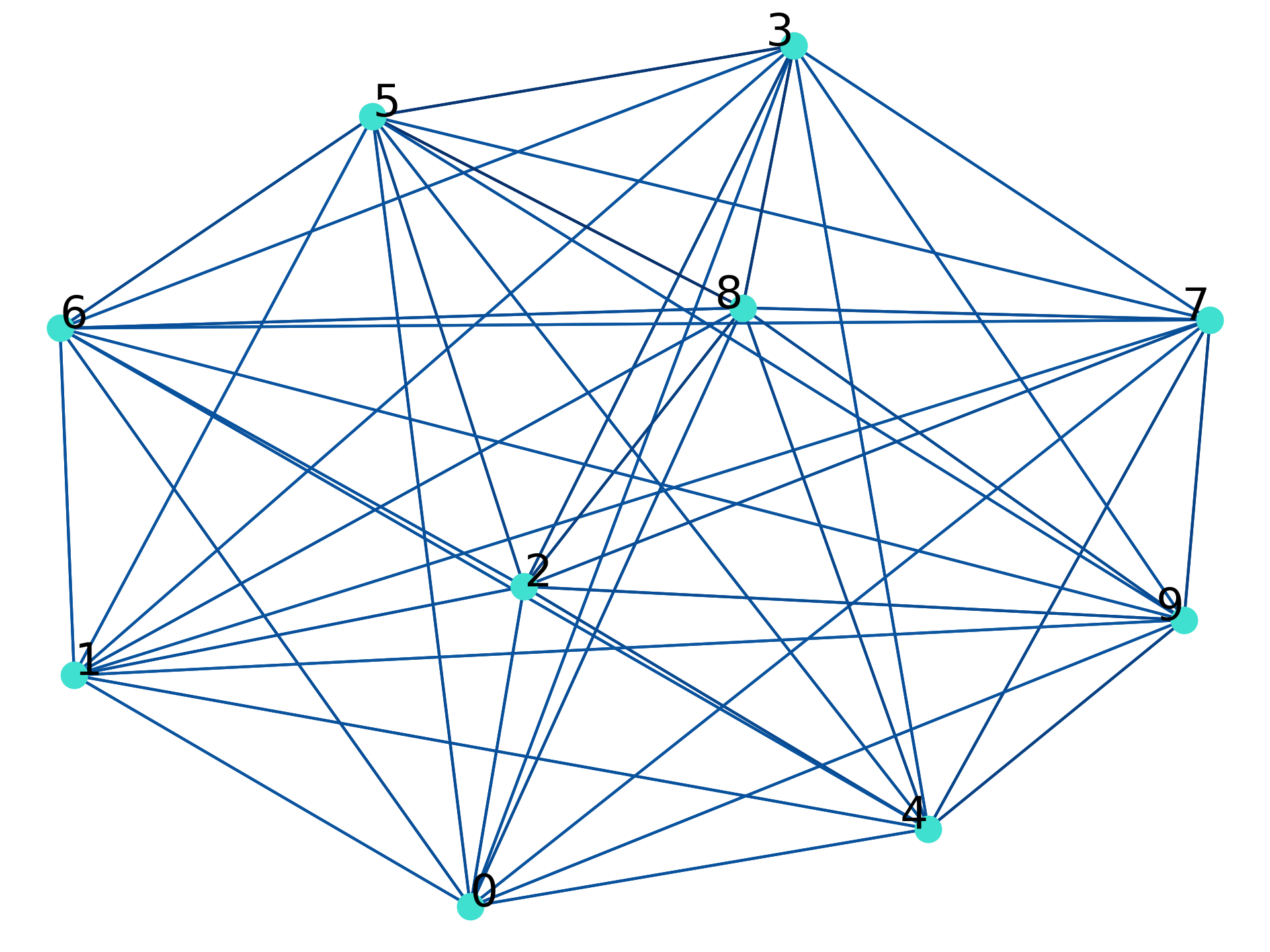} $\;\;\;\;\;$
\includegraphics[width=0.3\textwidth]{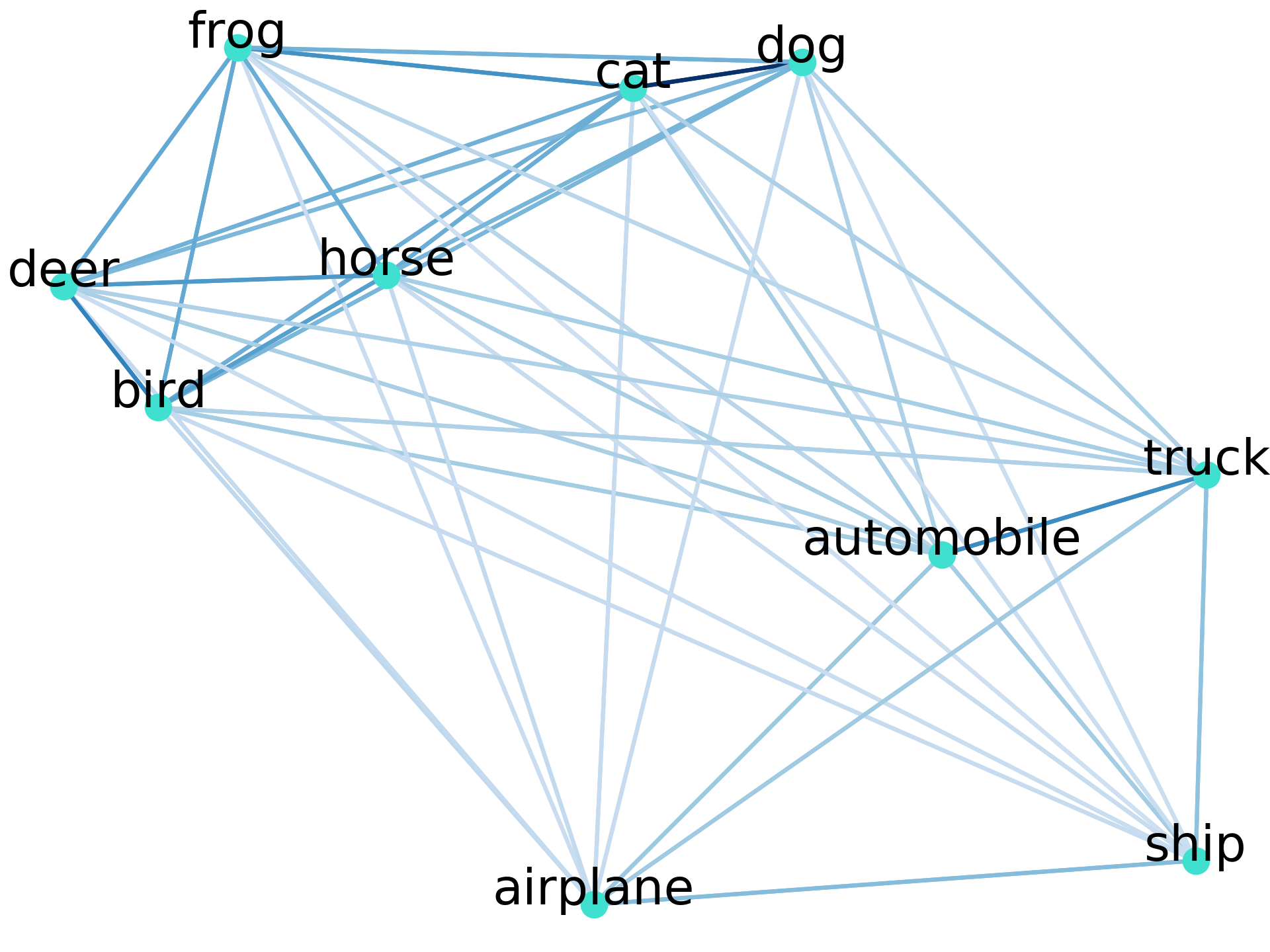} $\;\;\;\;\;$
\includegraphics[width=0.3\textwidth]{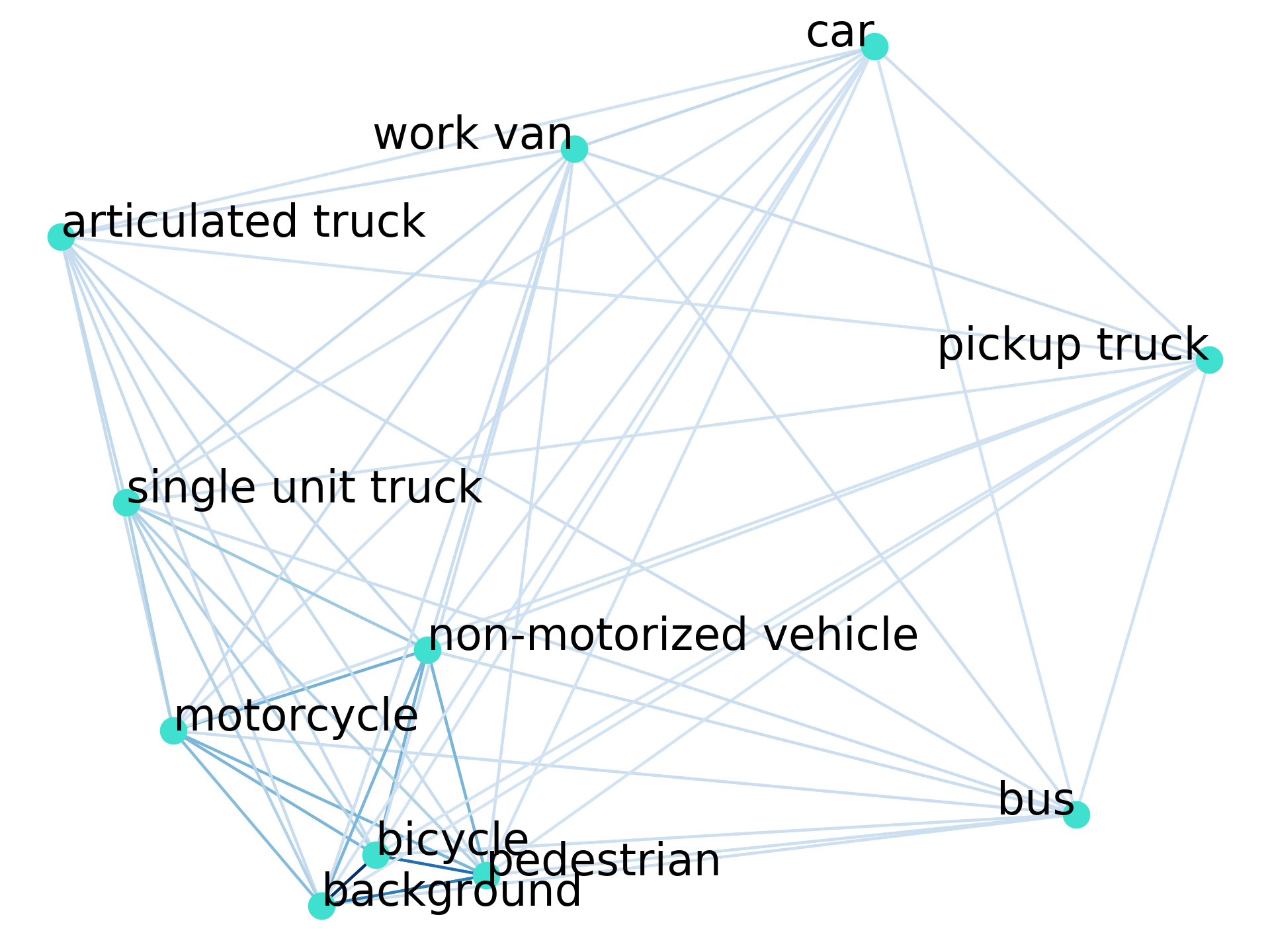}
\caption{2D plots of our W matrix for MNIST, CIFAR10 and MioTCD.}
\label{fig:MDS}
\end{figure*}

Dataset reduction can also be used to measure the similarity between two datasets with very different sizes like CIFAR10 (5,000 training samples per class) and STL-10 (500 training samples per class). While these datasets have visually similar content, the CNN error rates on it are very different (see Table~\ref{tab:datasets}). To measure the true distance between those datasets,  
we progressively reduced the number of samples for each  CIFAR10 class to reach that of STL-10. The results in Table~\ref{tab:reduce_cifar} show the close bound between our metric and the number of samples in the dataset. With only 500 samples, CIFAR10 gets a CSG score and a CNN accuracy similar but not identical to that of STL-10.  This shows that the datasets are similar but not identical, probably due to the fact that the CIFAR10 {\em Frog} class has been replaced by a {\em Monkey} class in STL-10 (supplementary material).  Here again, it took roughly one minute to produce the CSG scores (after having trained the embedding)  and 4 days for the CNN error rates.

\begin{figure}[h!]
    \centering
        \includegraphics[width=0.35\textwidth]{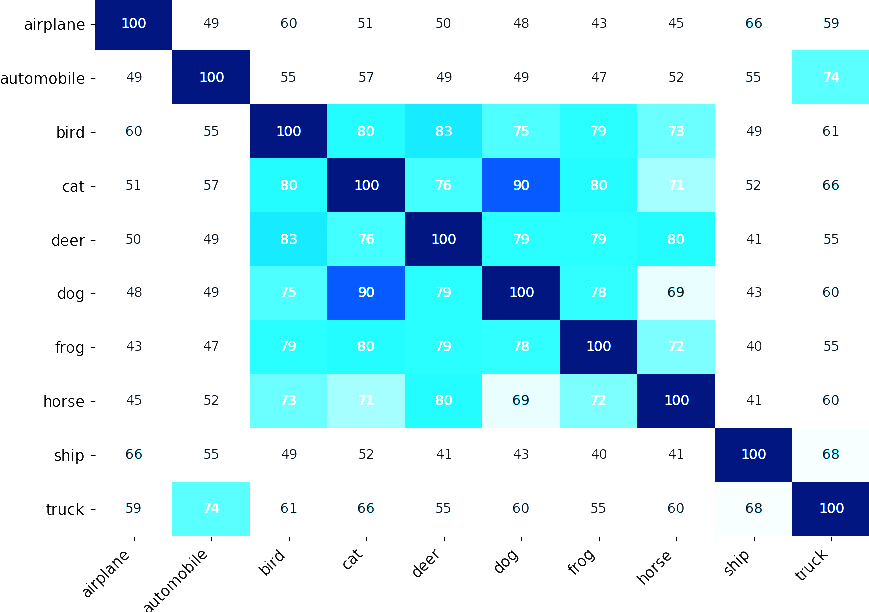}
        \includegraphics[width=0.35\textwidth]{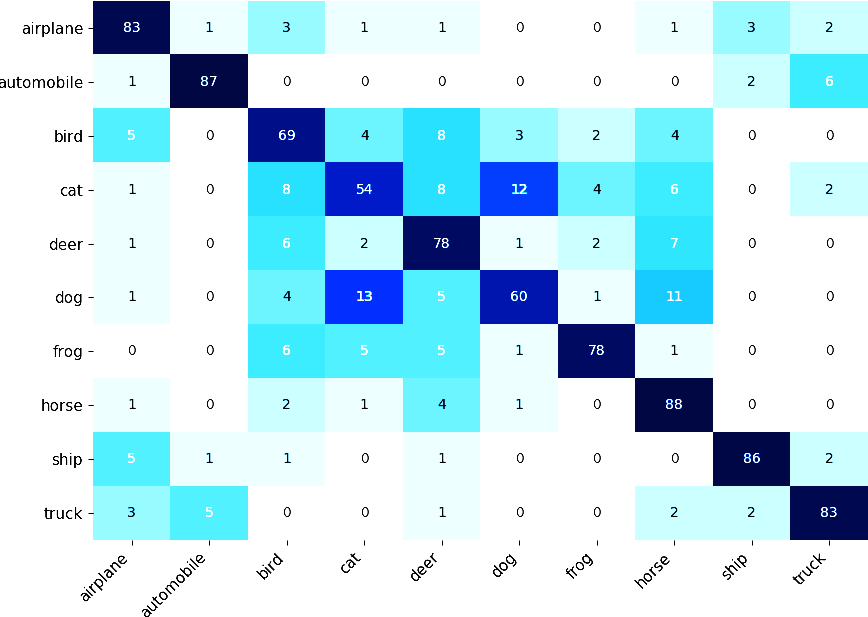}
\caption{[Top] our $W$ matrix and [Bottom] AlexNet's confusion matrix for CIFAR10. \vspace{-0.8cm}}
\label{fig:csg_matrix}
\end{figure}

\subsection{Confusion matrix}

While our c-measure can gauge the overall complexity of a dataset with a single measure that correlates with CNN accuracies, we can also use the similarity matrix $W$ (Eq.(\ref{eq:wij})) to analyze the inter-class distances.  As such, one can use a dissimilarity matrix $S=1-W$ to visualize the dataset in 2D via an algorithm such as multidimensional scaling (MDS)~\cite{borg2005modern}.  This results into plots such as those in Fig.~\ref{fig:MDS}.  While the classes of MNIST are all well separated, the CIFAR10 plots show that the {\em cat} and {\em dog} classes as well as {\em deer} and {\em bird} are close to each other, probably due to similar contexts.   As for MioTCD, the {\em bicycle, motorcycle} and {\em pedestrian} classes are in the same vicinity, mainly because of their small image resolution, they often contain more compression artifacts and hence be less feature-rich than other classes, making them confusing with featureless background.

As shown in Fig.~\ref{fig:csg_matrix}, the $W$ matrix strongly correlates to a real confusion matrix (here AlexNet). Here again, {\em cat} and {\em dog} as well as {\em deer} and {\em bird} are easily confused.   

\begin{table}[!h]
\centering
        {\footnotesize
        \begin{tabular}{|c|c|c|} \hline
        Dataset & CSG & Error rate \\ \hline
        CIFAR10 & 1.10 & 0.18 \\
        CIFAR10 reduced=4500 & 1.10 & 0.19 \\
        CIFAR10 reduced=3500 & 1.26 & 0.20 \\
        CIFAR10 reduced=2500 & 1.44 & 0.24 \\
        CIFAR10 reduced=1500 & 2.16 & 0.28 \\
        CIFAR10 reduced=500 & 2.59 & 0.42 \\
        STL-10 & 3.16 & 0.68 \\ \hline
        \end{tabular}}
        \caption{Effect of reducing the number of samples per class for CIFAR10 on our CSG metric and the AlexNet test error rate.}
        \label{tab:reduce_cifar}
\end{table}

\section{Conclusion}
In this work, we proposed a novel complexity measure designed for image classification problems called the cumulative spectral gradient (CSG) which is more accurate and faster than previous methods. We showed that our metric has many uses such as instance selection and class disentanglement. We also showed that the CSG closely matches the accuracy achievable by standard CNN architectures, an important feature when assessing an image dataset.

A future direction of our research would be to determine a procedure to compare the relative complexity of classification problems with different number of classes.  The analysis of random subsets of classes could be used as a common representation.  Another important direction would be to generalize our method to segmentation and localization problems.  As of now, it is not clear how these problems can be described by spectral clustering. 

Another future work would be to incorporate our similarity matrix $W$ in the optimization process of a neural network to minimize the interclass divergence. It is our intuition that the {\em a priori} knowledge of the interclass overlap could be used to force the optimizer to further separate entangled classes, a bit like the triplet loss does.
Finally, our metric is not restricted to image classification datasets and could be used in other areas of machine learning such as speech recognition and natural language processing (NLP). These fields already use state-of-the-art embeddings such as Word2Vec~\cite{Mikolov2013word2vec} and would thus naturally fall into our CSG framework.


\begin{thebibliography}{10}\itemsep=-1pt

\bibitem{abadi2016tensorflow}
M.~Abadi, P.~Barham, J.~Chen, Z.~Chen, A.~Davis, J.~Dean, M.~Devin,
  S.~Ghemawat, G.~Irving, M.~Isard, et~al.
\newblock Tensorflow: A system for large-scale machine learning.
\newblock In {\em OSDI}, volume~16, pages 265--283, 2016.

\bibitem{Anwar14}
N.~Anwar, G.~Jones, and S.~Ganesh.
\newblock Measurement of data complexity for classification problems with
  unbalanced data.
\newblock {\em Journal Stat. Anal. Data Mining}, 7(3):196--211, 2014.

\bibitem{Baumgartner06}
R.~Baumgartner and R.~Somorjai.
\newblock Data complexity assessment in undersampled classification of
  high-dimensional biomedical data.
\newblock {\em Pat. Rec. Letters}, 27(12), 2006.

\bibitem{becker2017seefood}
D.~Becker.
\newblock Hot dog - not hot dog.
\newblock [Online]. Available:
  url{https://www.kaggle.com/dansbecker/hot-dog-not-hot-dog}, 2017.

\bibitem{borg2005modern}
I.~Borg and P.~Groenen.
\newblock {\em Modern multidimensional scaling: theory and applications},
  volume~40.
\newblock Wiley Online Library, 2003.

\bibitem{bossard14food101}
L.~Bossard, M.~Guillaumin, and L.~{Van Gool}.
\newblock Food-101 -- mining discriminative components with random forests.
\newblock In {\em Proc. ECCV}, 2014.

\bibitem{Brun18}
A.~L. Brun, A.~S. Britto~Jr, L.~S. Oliveira, F.~Enembreck, and R.~Sabourin.
\newblock A framework for dynamic classifier selection oriented by the
  classification problem difficulty.
\newblock {\em Pat. Rec.}, 76:175--190, 2018.

\bibitem{bulatovNotMNIST}
Y.~Bulatov.
\newblock notmnist.
\newblock [Online]. Available: \url{www.kaggle.com/lubaroli/notmnist}, 2017.

\bibitem{chollet2016xception}
F.~Chollet.
\newblock Xception: Deep learning with depthwise separable convolutions.
\newblock In {\em Proc. CVPR}, pages 1800--1807, 2017.

\bibitem{chollet2015keras}
F.~Chollet et~al.
\newblock Keras.
\newblock [Online] Available :\url{https://github.com/fchollet/keras}, 2015.

\bibitem{coates2011analysis}
A.~Coates, A.~Ng, and H.~Lee.
\newblock An analysis of single-layer networks in unsupervised feature
  learning.
\newblock In {\em Proc. AISTATS}, pages 215--223, 2011.

\bibitem{Cummins11}
L.~Cummins and D.~Bridge.
\newblock Choosing a case base maintenance algorithm using a meta-case base.
\newblock In {\em Research and Development in Intelligent Systems XXVIII},
  pages 167--180. Springer, 2011.

\bibitem{dalal2005inria}
N.~Dalal and B.~Triggs.
\newblock Inria person dataset.
\newblock {\em [Online]. Available: http://pascal. inrialpes. fr/data/human},
  2005.

\bibitem{Duin06}
R.~Duin and E.~Pekalska.
\newblock {\em Object representation, sample size and dataset complexity}.
\newblock Springer, 2006.

\bibitem{Garcia15}
L.~Garcia, A.~de~Carvalho, and A.~Lorena.
\newblock Effect of label noise in the complexity of classification problems.
\newblock {\em Neurocomputing}, 160:108–119, 2015.

\bibitem{greenacre2013measures}
M.~Greenacre and R.~Primicerio.
\newblock Measures of distance between samples: non-euclidean.
\newblock {\em Multivariate analysis of ecological data}, pages 5--1, 2013.

\bibitem{he2016deep}
K.~He, X.~Zhang, S.~Ren, and J.~Sun.
\newblock Deep residual learning for image recognition.
\newblock In {\em Proc. CVPR}, pages 770--778, 2016.

\bibitem{ho2002complexity}
T.~Ho and M.~Basu.
\newblock Complexity measures of supervised classification problems.
\newblock {\em IEEE trans on PAMI}, 24(3):289--300, 2002.

\bibitem{hoiem2012diagnosing}
D.~Hoiem, Y.~Chodpathumwan, and Q.~Dai.
\newblock Diagnosing error in object detectors.
\newblock Springer, 2012.

\bibitem{ioffe2015batch}
S.~Ioffe and C.~Szegedy.
\newblock Batch normalization: Accelerating deep network training by reducing
  internal covariate shift.
\newblock In {\em Proc. ICML}, pages 448--456, 2015.

\bibitem{jaeger2014two}
S.~Jaeger, S.~Candemir, S.~Antani, Y.-X.~J. W{\'a}ng, P.-X. Lu, and G.~Thoma.
\newblock Two public chest x-ray datasets for computer-aided screening of
  pulmonary diseases.
\newblock {\em Quantitative imaging in medicine and surgery}, 4(6):475, 2014.

\bibitem{jebara2004probability}
T.~Jebara, R.~Kondor, and A.~Howard.
\newblock Probability product kernels.
\newblock {\em JMLR}, 5(7):819--844, 2004.

\bibitem{krizhevsky2009learning}
A.~Krizhevsky and G.~Hinton.
\newblock Learning multiple layers of features from tiny images.
\newblock {\em Tech.Report}, 2009.

\bibitem{krizhevsky2012imagenet}
A.~Krizhevsky, I.~Sutskever, and G.~E. Hinton.
\newblock Imagenet classification with deep convolutional neural networks.
\newblock In {\em Proc. NIPS}, pages 1097--1105, 2012.

\bibitem{lecun2010mnist}
Y.~LeCun, C.~Cortes, and C.~Burges.
\newblock Mnist handwritten digit database.
\newblock {\em AT\&T Labs [Online]. Available: http://yann. lecun.
  com/exdb/mnist}, 2, 2010.

\bibitem{Leyva15}
E.~Leyva, A.~González, and R.~Pérez.
\newblock A set of complexity measures designed for applying meta-learning to
  instance selection.
\newblock {\em IEEE Trans on Knowledge and Data Eng.}, 27(2):354--367, 2015.

\bibitem{li2018measuring}
C.~Li, H.~Farkhoor, R.~Liu, and J.~Yosinski.
\newblock Measuring the intrinsic dimension of objective landscapes.
\newblock In {\em Proc. ICLR}, 2018.

\bibitem{Lorena18}
A.~{Lorena}, L.~{Garcia}, J.~{Lehmann}, M.~{Souto}, and T.~{Ho}.
\newblock {How Complex is your classification problem? A survey on measuring
  classification complexity}.
\newblock {\em ArXiv --1808.03591v1}, 2018.

\bibitem{Luo18}
Z.~Luo, F.~{B.Charron}, C.~Lemaire, J.~Konrad, S.~Li, A.~Mishra, A.~Achkar,
  J.~Eichel, and P.-M. Jodoin.
\newblock Mio-tcd: A new benchmark dataset for vehicle classification and
  localization.
\newblock {\em In press at IEEE TIP}, 2018.

\bibitem{van2008visualizing}
L.~v.~d. Maaten and G.~Hinton.
\newblock Visualizing data using t-sne.
\newblock {\em JMLR}, 9(11):2579--2605, 2008.

\bibitem{Mantovani15}
R.~G. Mantovani, A.~L. Rossi, J.~Vanschoren, B.~Bischl, and A.~C. Carvalho.
\newblock To tune or not to tune: recommending when to adjust svm
  hyper-parameters via meta-learning.
\newblock In {\em Proc. IJCNN}, 2015.

\bibitem{Mikolov2013word2vec}
T.~Mikolov, I.~Sutskever, K.~Chen, G.~Corrado, and J.~Dean.
\newblock Distributed representations of words and phrases and their
  compositionality.
\newblock In {\em Proc. NIPS}, 2013.

\bibitem{Mohar1997}
B.~Mohar.
\newblock {\em Some applications of Laplace eigenvalues of graphs}, pages
  225--275.
\newblock Springer, 1997.

\bibitem{morais2013complex}
G.~Morais and R.~C. Prati.
\newblock Complex network measures for data set characterization.
\newblock In {\em Proc. BRACIS}, pages 12--18. IEEE, 2013.

\bibitem{netzer2011reading}
Y.~Netzer, T.~Wang, A.~Coates, A.~Bissacco, B.~Wu, and A.~Ng.
\newblock Reading digits in natural images with unsupervised feature learning.
\newblock In {\em Proc. NIPS}, page~5, 2011.

\bibitem{Nowakowska14}
E.~{Nowakowska}, J.~{Koronacki}, and S.~{Lipovetsky}.
\newblock {Tractable Measure of Component Overlap for Gaussian Mixture Models}.
\newblock {\em ArXiv -- 1407.7172}, 2014.

\bibitem{orriols2010documentation}
A.~Orriols-Puig, N.~Macia, and T.~K. Ho.
\newblock Documentation for the data complexity library in c++.
\newblock {\em Universitat Ramon Llull, La Salle}, 196, 2010.

\bibitem{Saez13}
J.~Saez, J.~Luengo, and F.~Herrera.
\newblock Predicting noise filtering efficacy with data complexity measures for
  nearest neighbor classification.
\newblock {\em Pat. Rec.}, 46(1):355--364, 2013.

\bibitem{simonyan2014very}
K.~Simonyan and A.~Zisserman.
\newblock Very deep convolutional networks for large-scale image recognition.
\newblock In {\em Proc. ACPR}, 2015.

\bibitem{Sorgun2013bounds}
S.~Sorgun and S.~Buyukkose.
\newblock Bounds for the largest laplacian eigenvalues of weighted graphs.
\newblock {\em International Journal of Combinatorics}, 2013, 2013.

\bibitem{Sotoca06}
J.~Sotoca, R.~Mollineda, and J.~Sanchez.
\newblock A meta-learning framework for pattern classification by means of data
  complexity measures.
\newblock {\em Int. art. rev. iber de IA}, 10(29):31--38, 2006.

\bibitem{von2007tutorial}
U.~Von~Luxburg.
\newblock A tutorial on spectral clustering.
\newblock {\em Statistics and computing}, 17(4):395--416, 2007.

\bibitem{Wang05}
L.~Wang, Y.~Zhang, and J.~Feng.
\newblock On the euclidean distance of images.
\newblock {\em IEEE Trans on PAMI}, 27(8), 2005.

\bibitem{yang2015large}
L.~Yang, P.~Luo, C.~Change~Loy, and X.~Tang.
\newblock A large-scale car dataset for fine-grained categorization and
  verification.
\newblock In {\em Proc. CVPR}, pages 3973--3981, 2015.

\bibitem{Zhang17}
C.~Zhang, S.~Bengio, M.~Hardt, B.~Recht, and O.~Vinyals.
\newblock Understanding deep learning requires re-thinking generalization.
\newblock In {\em Proc. ICLR}, 2017.

\end{thebibliography}
\end{document}


%
\title{Supplementary Material\\ Spectral Metric for Dataset Complexity Assessment\vspace{-1ex}}
\maketitle

$\,$

\section{Instance selection}
\begin{figure}[t]
\centering
\includegraphics[width=0.45\textwidth]{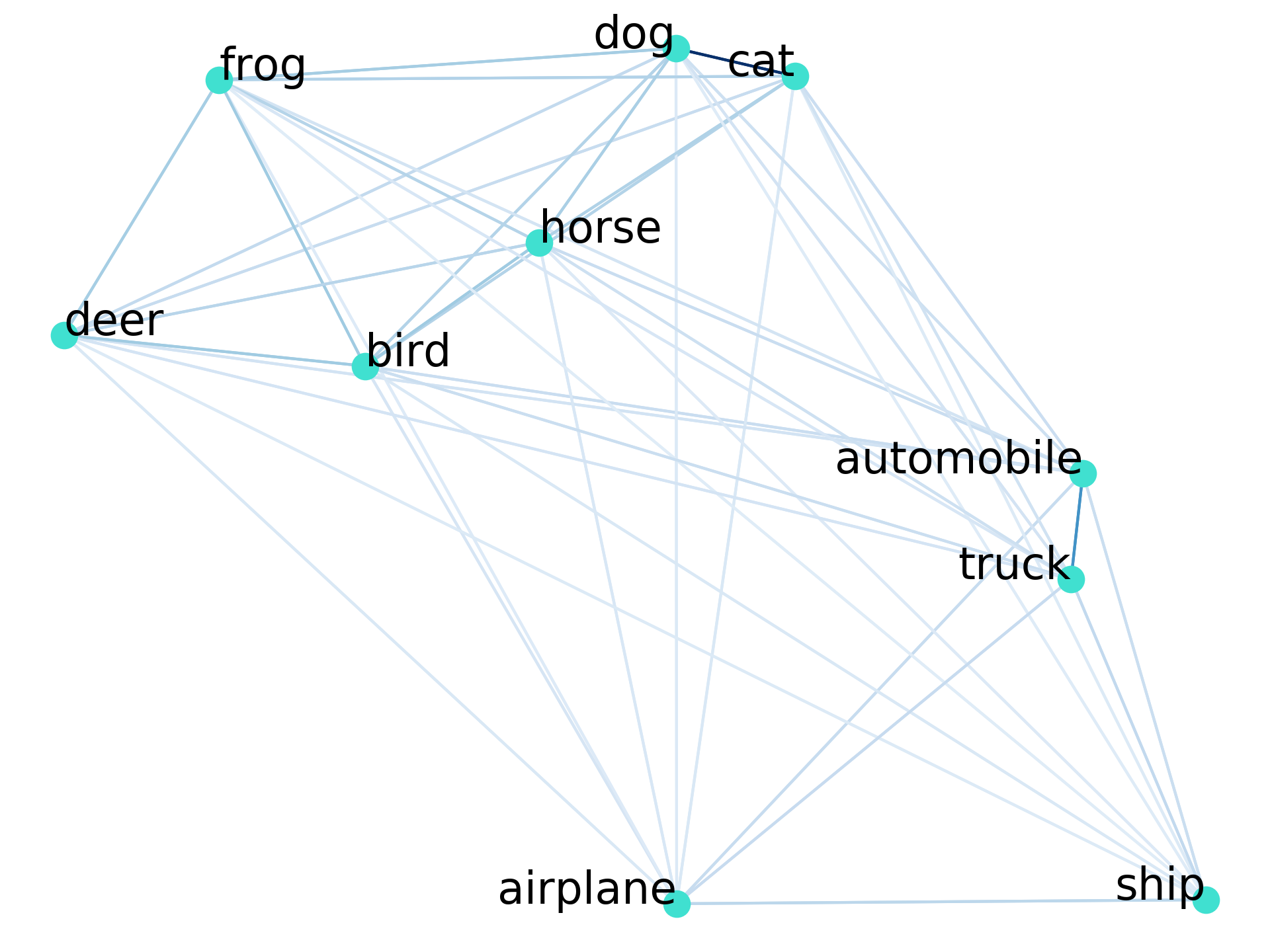}
\includegraphics[width=0.45\textwidth]{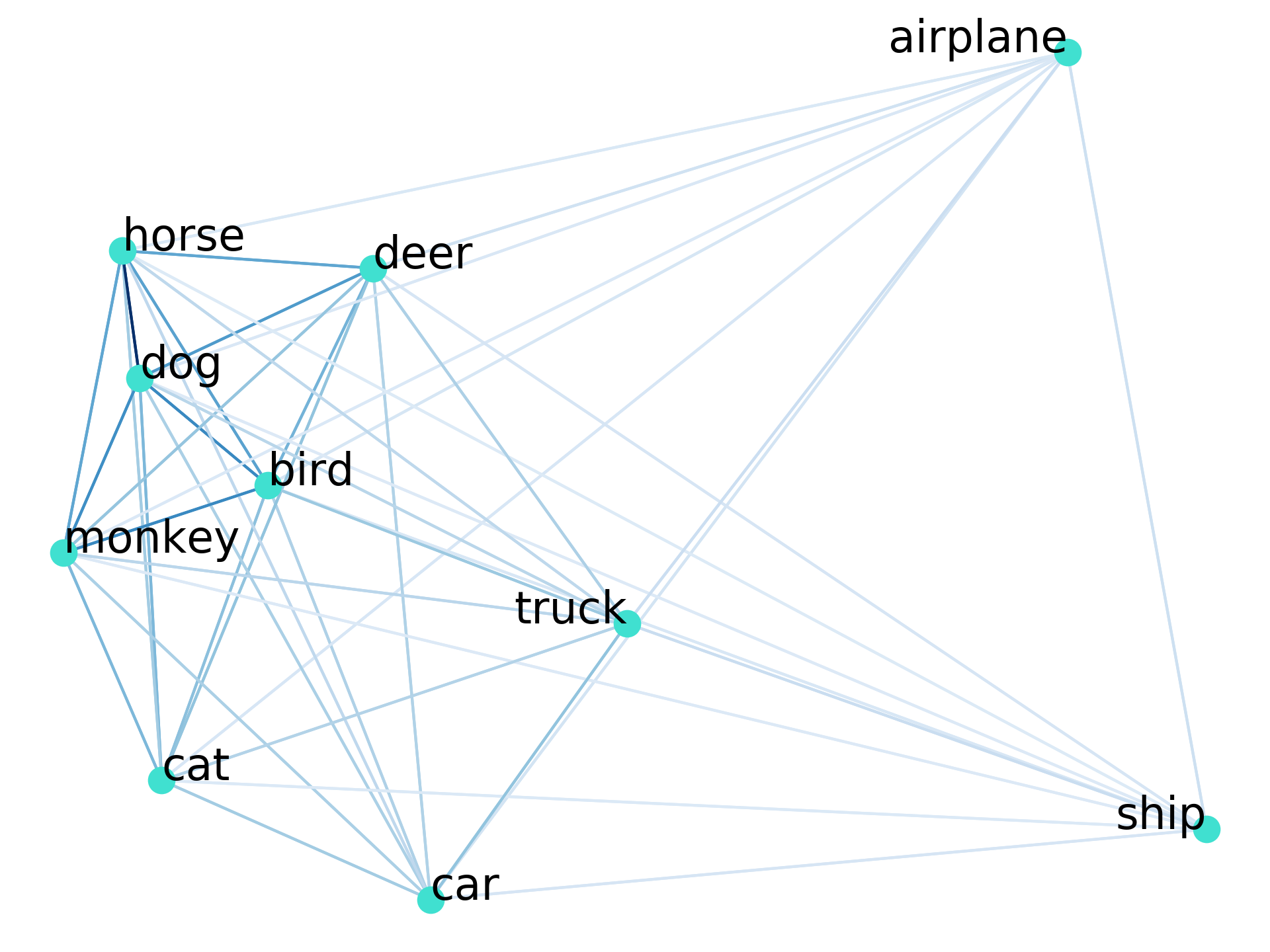}
\caption{Comparison between the [Top] 2D plot of CIFAR10 with 500 samples per class and [Bottom]  2D plot of STL-10. As we see, the nodes in STL-10 seems closer than in CIFAR10.}
\label{fig:MDS_2}
\end{figure}

In the results section, we showed that STL-10 was a bit harder than CIFAR10 with 500 samples per class even though these datasets are visually very similar. To further our point, we show in Fig.~\ref{fig:MDS_2} the MDS representation of both datasets. While the {\em car-truck} distance is smaller in CIFAR10, the six animal classes in STL-10 are overall closer together and thus a slightly more difficult to disentangle.



\section{Comparison with other graph-based methods}
We mentioned in the {\em Previous works} section that other graph-based methods have been proposed in the past. One could wonder how our measure differs from those. The main advantage of our approach compared to other graph-based methods is the fact that our graph embeds classes and not samples, thanks to the spectral clustering formalism.  This leads to a $K\times K$ Laplacian matrix which is order of magnitude smaller that the  $N \times N$ similarity matrix often required by other methods. This brings a huge advantage both memory and processing wise while allowing our method to naturally expand to the number of classes.  

\section{How to choose the auto-encoder architecture?}

Empirical evidences show that any auto-encoder that properly reconstruct images can be used with our method.  The key element for our method is to have low-dimensional space which correctly regroup images with similar content.

\begin{table*}[]
    \centering
    \begin{tabular}{c|ccccccccccccc}
$CNN_{AE}$ & CSG & F1 & F2 & F3 & F4 & N1 & N2 & N3 & N4 & T1 & T2 & AlexNet Acc. & Resnet Acc. \\
\hline
compcars & 6.421 & 0.025 & 2.10E-15 & 0.031 & 3.582 & 0.796 & 1.018 & 0.668 & 0.176 & 1 & 23.27 & 0.303 & 0.122 \\
stl10 & 5.051 & 0.243 & 2.56E-15 & 0.056 & 4.915 & 0.869 & 1.075 & 0.78 & 0.609 & 1 & 19.531 & 0.314 & 0.367 \\
cifar10 & 3.509 & 0.331 & 1.37E-14 & 0.022 & 0.87 & 0.712 & 1.002 & 0.618 & 0.468 & 1 & 195.312 & 0.821 & 0.807 \\
svhn & 3.986 & 0.008 & 1.28E-14 & 0.01 & 0.406 & 0.637 & 0.989 & 0.525 & 0.646 & 1 & 286.16 & 0.918 & 0.933 \\
notMNIST & 0.564 & 0.756 & 1.30E-11 & 0.133 & 6.915 & 0.226 & 0.599 & 0.155 & 0.042 & 1 & 73.141 & 0.951 & 0.96 \\
mnist & 0.202 & 0.723 & 2.44E-19 & 0.258 & 2.803 & 0.078 & 0.7 & 0.039 & 0.021 & 0.998 & 234.375 & 0.994 & 0.954 \\
\end{tabular}

    \caption{Complexity estimation for all 10 class datasets for all methods using $CNN_{AE}$ embedding.}
    \label{tab:cnn}
\end{table*}

\begin{table*}[]
    \centering
    \begin{tabular}{c|ccccccccccccc}
CNN-TSNE & CSG & F1 & F2 & F3 & F4 & N1 & N2 & N3 & N4 & T1 & T2 & AlexNet Acc. & Resnet Acc. \\
\hline
compcars & 2.928 & 0.014 & 35.578 & 0.031 & 0.084 & 0.845 & 1.633 & 0.632 & 0.874 & 0.993 & 1985.667 & 0.303 & 0.122 \\
stl10 & 3.072 & 0.222 & 37.264 & 0.031 & 0.073 & 0.91 & 2.426 & 0.788 & 0.831 & 0.999 & 1666.667 & 0.314 & 0.367 \\
cifar10 & 1.004 & 0.291 & 36.344 & 0.009 & 0.017 & 0.797 & 1.585 & 0.622 & 0.819 & 0.998 & 16666.666 & 0.821 & 0.807 \\
svhn & 1.154 & 0.001 & 42.921 & 0.002 & 0.004 & 0.748 & 1.408 & 0.546 & 0.89 & 0.999 & 24419 & 0.918 & 0.933 \\
notMNIST & 0.725 & 1.752 & 33.723 & 0.083 & 0.212 & 0.234 & 0.284 & 0.158 & 0.434 & 0.999 & 6241.333 & 0.951 & 0.96 \\
mnist & 0.112 & 6.193 & 26.706 & 0.229 & 0.522 & 0.055 & 0.112 & 0.03 & 0.104 & 0.982 & 20000 & 0.994 & 0.954 \\
\end{tabular}

    \caption{Complexity estimation for all 10 class datasets for all methods using $CNN_{AE}$ t-SNE embedding.}
    \label{tab:cnntsne}
\end{table*}

\begin{table*}[]
    \centering
    \begin{tabular}{c|ccccccccccccc}
t-SNE & CSG & F1 & F2 & F3 & F4 & N1 & N2 & N3 & N4 & T1 & T2 & AlexNet Acc. & Resnet Acc. \\
\hline
compcars & 2.079 & 0.018 & 30.066 & 0.03 & 0.092 & 0.857 & 1.518 & 0.675 & 0.867 & 0.99 & 1191.4 & 0.303 & 0.122 \\
stl10 & 1.761 & 0.822 & 20.818 & 0.199 & 0.415 & 0.843 & 1.512 & 0.704 & 0.756 & 0.999 & 1000 & 0.314 & 0.367 \\
cifar10 & 0.858 & 0.294 & 32.037 & 0.012 & 0.032 & 0.805 & 1.504 & 0.643 & 0.816 & 0.996 & 10000 & 0.821 & 0.807 \\
svhn & 0.967 & 0.007 & 38.829 & 0.009 & 0.02 & 0.712 & 1.265 & 0.518 & 0.88 & 1 & 14651.4 & 0.918 & 0.933 \\
notMNIST & 0.709 & 1.847 & 24.405 & 0.197 & 0.513 & 0.193 & 0.288 & 0.136 & 0.356 & 0.999 & 3744.8 & 0.951 & 0.96 \\
mnist & 0.151 & 4.721 & 16.875 & 0.393 & 1.112 & 0.051 & 0.105 & 0.029 & 0.083 & 0.998 & 12000 & 0.994 & 0.954 \\
\end{tabular}

    \caption{Complexity estimation for all 10 class datasets for all methods using t-SNE embedding.}
    \label{tab:tsne}
\end{table*}

\begin{table*}[]
    \centering
    \begin{tabular}{c|ccccccccccccc}
Raw & CSG & F1 & F2 & F3 & F4 & N1 & N2 & N3 & N4 & T1 & T2 & AlexNet Acc. & Resnet Acc. \\
\hline
compcars & 6.700 & 0.031 & 2.10E-43 & 0.053 & 5.962 & 0.81 & 1.024 & 0.713 & 0.172 & 0.99 & 0.121 & 0.618 & 0.52 \\
stl10 & 3.349 & 0.324 & 7.81E-06 & 0.243 & 16.97 & 0.795 & 1.037 & 0.732 & 0.418 & 0.99 & 0.181 & 0.994 & 0.954 \\
cifar10 & 3.579 & 0.244 & 3.54E+01 & 0.008 & 0.975 & 0.742 & 1.012 & 0.649 & 0.602 & 0.99 & 16.276 & 0.951 & 0.96 \\
svhn & 3.889 & 0.005 & 2.45E-06 & 0.004 & 3 & 0.626 & 0.987 & 0.511 & 0.667 & 0.99 & 23.847 & 0.897 & 0.896 \\
notMNIST & 0.512 & 0.996 & 4.50E+01 & 0 & 0 & 0.198 & 0.577 & 0.137 & 0.038 & 1 & 7.961 & 0.918 & 0.933 \\
mnist & 0.084 & 0.735 & 0.00E+00 & 0.243 & 5.203 & 0.054 & 0.7 & 0.026 & 0.007 & 0.99 & 76.531 & 0.314 & 0.367 \\
\end{tabular}

    \caption{Complexity estimation for all 10 class datasets for all methods using no embedding.}
    \label{tab:raw}
\end{table*}

\begin{table*}[ht!]
    \centering
    {\small
    \begin{tabular}{c|C{0.8cm}C{0.8cm}C{0.8cm}C{1cm}}
         & \multicolumn{4}{c}{Alexnet} \\ \hline
         Method & Raw & {\em CNN{\tiny $_{AE}$}} & {\em t-SNE} & {\em CNN{\tiny $_{AE}$} t-SNE} \\
        \hline
        N4 & 0.141 & 0.558 & 0.763 & 0.780  \\
        F3 & 0.290 & 0.665 & 0.459 & 0.604  \\
        F1 & 0.483 & 0.895 & 0.667 & 0.673  \\
        F2 & 0.366 & 0.449 & 0.234 & 0.660  \\
        T1 & 0.642 & 0.519 & 0.505 & 0.332  \\
        T2 & 0.655 & 0.783 & 0.783 & 0.783  \\
        N2 & 0.677 & 0.812 & 0.794 & 0.854  \\
        F4 & 0.760 & 0.067 & 0.531 & 0.606  \\
        N1 & 0.767 & 0.861 & 0.817 & 0.833  \\
        N3 & 0.803 & 0.855 & 0.823 & 0.837  \\
        \hline
        Our Method & 0.696 & 0.823 & 0.903 & \textbf{0.968}\\
    \end{tabular}
        \begin{tabular}{||C{0.8cm}C{0.8cm}C{0.8cm}C{1cm}}
          \multicolumn{4}{c}{ResNet} \\ \hline
         Raw & {\em CNN{\tiny $_{AE}$}} & {\em t-SNE} & {\em CNN{\tiny $_{AE}$} t-SNE} \\
        \hline
         0.021 & 0.407 & 0.746 & 0.745  \\
         0.201 & 0.637 & 0.500 & 0.552  \\
         0.519 & 0.922 & 0.648 & 0.626  \\
        0.432 & 0.470 & 0.320 & 0.576  \\
         0.642 & 0.461 & 0.677 & 0.214  \\
         0.578 & 0.737 & 0.737 & 0.737  \\
         0.619 & 0.775 & 0.761 & 0.760  \\
         0.606 & 0.001 & 0.542 & 0.551  \\
         0.700 & 0.809 & 0.784 & 0.783  \\
        0.727 & 0.794 & 0.781 & 0.765  \\
        \hline
         0.712 & 0.838 & 0.932 & \textbf{0.935}\\
    \end{tabular}
            \begin{tabular}{||C{0.8cm}C{0.8cm}C{0.8cm}C{1cm}}
          \multicolumn{4}{c}{XceptionNet} \\ \hline
          Raw & {\em CNN{\tiny $_{AE}$}} & {\em t-SNE} & {\em CNN{\tiny $_{AE}$} t-SNE} \\
        \hline
         0.067 & 0.558 & 0.763 & 0.780 \\
        0.267 & 0.665 & 0.459 & 0.604 \\
         0.458 & 0.895 & 0.669 & 0.673 \\
         0.465 & 0.449 & 0.234 & 0.660 \\
         0.642 & 0.519 & 0.505 & 0.332 \\
         0.578 & 0.783 & 0.783 & 0.783 \\
         0.564 & 0.816 & 0.794 & 0.854 \\
         0.676 & 0.067 & 0.531 & 0.606 \\
         0.651 & 0.861 & 0.817 & 0.833 \\
         0.681 & 0.855 & 0.823 & 0.837 \\
        \hline
         0.718 & 0.804 & 0.931 & \textbf{0.951}\\
    \end{tabular}
    }
    
    \caption{Correlation values between the accuracy of AlexNet, ResNet and XceptionNet and  Ho \& Basu c-measures with four different embeddings.  These results complement Table 3, 4 and 5 in the original paper.\vspace{-0.6 cm}}
    \label{tab:corr}
\end{table*}